# A Method for Modeling Co-Occurrence Propensity of Clinical Codes with Application to ICD-10-PCS Auto-Coding

Michael Subotin, Anthony R. Davis
3M Health Information Systems

**ABSTRACT**

**Objective.** Natural language processing methods for medical *auto-coding,* or automatic generation of medical billing codes from electronic health records, generally assign each code independently of the others. They may thus assign codes for closely related procedures or diagnoses to the same document, even when they do not tend to occur together in practice, simply because the right choice can be difficult to infer from the clinical narrative.

**Materials and Methods.** We propose a method that injects awareness of the propensities for code co-occurrence into this process. First, a model is trained to estimate the conditional probability that one code is assigned by a human coder, given than another code is known to have been assigned to the same document. Then, at runtime, an iterative algorithm is used to apply this model to the output of an existing statistical auto-coder to modify the confidence scores of the codes.

**Results**. We tested this method in combination with a primary auto-coder for ICD-10 procedure codes, achieving a 12% relative improvement in F-score over the primary auto-coder baseline.

**Discussion**. The proposed method can be used, with appropriate features, in combination with any auto-coder that generates codes with different levels of confidence.

**Conclusion**. The promising results obtained for ICD-10 procedure codes suggest that the proposed method may have wider applications in auto-coding.

## BACKGROUND AND SIGNIFICANCE

### Introduction

In many countries, reimbursement rules for health care services stipulate that clinical documentation of patient care must be assigned codes for the diagnoses and procedures described therein. These codes may be assigned by general health care personnel or by specially trained medical coders. The billing codes used in the United States include International Statistical Classification of Diseases and Related Health Problems (ICD) codes, whose version 9 is currently in use and whose version 10 is scheduled for adoption in October 2015, as well as Current Procedural Terminology (CPT) codes. The same codes are also used for research, internal bookkeeping, and other purposes.

Assigning codes to clinical documentation often requires extensive technical training and involves substantial labor costs. This, together with increasing prominence of electronic health records (EHRs), has prompted development and adoption of natural language processing (NLP) algorithms that support the coding workflow by automatically inferring appropriate codes from the clinical narrative and other information contained in the EHR [1-5]. The need for innovation in computer-assisted coding becomes especially acute with the introduction of ICD-10 and the associated increase of training and labor costs for manual coding. The novelty and complexity of ICD-10 presents unprecedented challenges for developers of rule-based auto-coding software. In turn, statistical auto-coding methods are constrained by the scarcity of available training data with human-assigned ICD-10 codes.

The present study concentrates on a particular challenge common to all auto-coding algorithms, regardless of the code set they work with: how to take into account relationships among codes generated for the same EHR. The simplest approach is to generate all the codes independently of one another, but this often leads to errors, since the practical restrictions on possible code combinations are ignored. For example, the ICD-10-PCS codes for hip replacement contain a character representing the material the prosthesis is made of (unspecified, metal, ceramic, etc.) In practice, only one of these options can apply in a given case, but an auto-coder that ignores relationships between codes can easily generate more than one of these codes, simply because the right choice can be difficult to infer from the clinical narrative. Cases like these could be handled by applying rule-based logic to the output of the auto-coder. However, other relationships between codes and the underlying clinical facts are less clear-cut. Thus, some procedures are often but not always performed together, while others could only appear in the same patient encounter by pure chance, even though their co-occurrence cannot be logically ruled out. These propensities cannot be reliably encoded by rules, but they are a natural fit for statistical methods. This is especially the case when the primary auto-coder itself utilizes statistical methods, so that the generated codes are supplied with confidence scores. These scores can then be combined with additional confidence scores that take into account the propensities of code co-occurrence for improved accuracy of predictions.

**ICD-10 Procedure Coding System (PCS)**

We test the proposed approach in application to ICD-10-PCS, a set of codes for medical procedures developed by 3M Health Information Systems under contract with the Center for Medicare and Medicaid Services of the U.S. government. ICD-10-PCS has been designed systematically; each code consists of seven characters, and the character in each of these positions signifies one particular aspect of the code. The first character designates the "section" of ICD-10-PCS: 0 for Medical and Surgical, 1 for Obstetrics, 2 for Placement, and so on. Within each section, the seven components, or axes of classification, are intended to have a consistent meaning; for example in the Medical and Surgical section, the second character designates the body system involved, the third the root operation, and so on (see Table 1). All procedures in this section are thus classified along these axes. For instance, in a code such as *0DBJ3ZZ*, *D* in the second position indicates that the body system involved is the gastrointestinal system, *B* in the third position always indicates that the root operation is an excision of a body part, *J* in the fourth position indicates that the appendix is the body part involved, and *3* in the fifth position indicates that the approach is percutaneous. The values *Z* in the last two axes mean that neither a device nor a qualifier is specified.

| Character | Meaning |
|---|---|
| 1st | Section |
| 2nd | Body System |
| 3rd | Root Operation |
| 4th | Body Part |
| 5th | Approach |
| 6th | Device |
| 7th | Qualifier |

Table 1. Character Specification of the Medical and Surgical Section of ICD-10-PCS.

Like its counterpart for diagnoses, ICD-10-PCS introduces many distinctions absent in ICD-9. As a result, it contains over 70,000 distinct codes, which presents challenges for both manual coding and auto-coding.

The systematic organization of ICD-10-PCS means that we can treat codes as combinations of features or concepts drawn from each of the dimensions defined for a section of the code set. We exploit the conceptual structure of ICD-10-PCS by means of a curated crosswalk from codes to a set of clinical concepts.

## OBJECTIVE

We present a model that infers code co-occurrence propensities from data, and furthermore does so by analyzing the internal structure of the codes. It is trained on a corpus of EHRs with manually assigned codes, where only the codes and not the rest of the EHRs are used for training. We also propose an algorithm to apply this model at run-time. It uses an approximate computational strategy called "greedy" in computer science. In their general form, the proposed model and run-time algorithm can be applied to any code set and any primary auto-coder that generates confidence scores. We illustrate their implementation with a detailed description of their application to the ICD-10 Procedure Coding System (PCS). In particular, we show how to parametrize the model so that it makes statistical generalizations based on attributes of the codes and produce probability estimates even for previously unseen code pairs.

Although we concentrate on co-occurrence propensities of codes of the same type, modeling cross-family co-occurrence of procedure and diagnosis codes can also be used to improve accuracy in an auto-coder that generates both types of codes, in essence supplying a probabilistic counterpart to medical necessity considerations. Thus, if the auto-coder predicts a procedure code but cannot identify any of the diagnoses that normally motivate it, this should prompt a decrease of confidence in that prediction. Because of the differences between the two types of codes, they would generally be predicted by distinct components, whose confidence scores may not be directly comparable, which presents additional challenges in combining the scores of the primary auto-coder with co-occurrence model scores. We discuss how these challenges can be tackled within the proposed framework.

## MATERIALS AND METHODS

### A Model Of Code Co-Occurrence Propensity

The proposed model of code co-occurrence estimates the probability, $P(C_i|C_j)$, that a code $C_i$ would be observed (i.e., appropriately assigned) given that the code $C_j$ is known to have been observed for the same EHR (which can refer to a single clinical note or documentation for an entire patient encounter, whichever is appropriate). Note that the model has only two possible outcomes for each computation: $C_i$ is *observed* and $C_i$ is *not observed*. It does not define a single probability distribution over all the possible codes, which would be much more computationally expensive to estimate at run-time. Note also that probabilities for all code pairs are estimates by a single model, rather than separate models for each $C_j$, which would lead to fragmentation of training data and prevent generalization of the kind described below.

A number of statistical methods can be used to estimate this probability. We use $\ell_1$-regularized logistic regression (also known as maximum entropy), which has shown good performance for NLP tasks in terms of accuracy as well as scalability at training and run-time [6]. Logistic regression can make use of features that track arbitrary aspects of the observation and the predicted outcome label.

This model is intended to capture solely the trends of code co-occurrence, leaving prediction of individual codes from the EHR to the primary auto-coder. Therefore, it does not use features that depend on the body of the EHR. We accordingly restrict the model features to observations of pairs of codes, and track their various aspects. In a later section we will illustrate, using ICD-10-PCS as an example, how to define these features so that the model can make generalizations for code co-occurrences or even codes that were not observed in its training data.

The algorithm for generation of training instances is shown in Figure 1. Its input is produced by running the primary auto-coder on a set of documents with manually assigned codes. The top-scoring codes generated by the auto-coder (selected by picking a fixed number of codes or setting a fixed threshold on the auto-coder's confidence score), together with manually assigned codes, become the candidate codes in the training instances, conditioned on one of the manually assigned codes. In practice, this generates many more negative than positive instances, which motivates sub-sampling the negative instances.

1. **Input**:
2.   $D_1 \ldots D_M$ : a set of $M$ documents with manually assigned codes
3.   $MAN(D_1) \ldots MAN(D_M)$: sets of manually assigned codes
4.   $GEN(D_1) \ldots GEN(D_M)$: top-scoring outputs of primary auto-coder

5. **For each** $D_i$ in $D_1 \ldots D_M$:
6.   **For each** $C_j^{man}$ in $MAN(D_i)$:
7.     **For each** $C_k^{pred}$ in $GEN(D_i) \cup MAN(D_i)$:
8.       Extract features for estimate $P(C_k^{pred}|C_j^{man})$
9.       **If** $C_k^{pred} \in MAN(D_i)$:
10.         Generate positive training instance
11.      **else**:
12.         Generate negative training instance

Figure 1. Algorithm for generation of training instances. An example of feature extraction logic is given below.

The algorithm restricts generation of training instances to possible code pairs found in the top-scoring outputs generated by the primary auto-coder. Thus, strictly speaking, the model $P(C_i|C_j)$ answers the following question: if the code $C_j$ is observed and the code $C_i$ has been generated by the primary auto-coder, what is the probability that the code $C_i$ is appropriate? This focuses the model on the actual errors that the given auto-coder is likely to make and which could be corrected by taking co-occurrence propensities into account.

**A Greedy Run-Time Algorithm**

A model trained as described in the previous section can estimate the probability of observing arbitrary codes, conditional on the knowledge that a given code has been appropriately assigned. At run-time we do not know in advance what codes are appropriate for the document, and hence we do not have the information to compute these conditional probabilities. In order to apply this model at run-time, we use the approximation described in Figure 2.

1. **Input**:
2.    $GEN(D_1) \dots GEN(D_M)$: top-scoring outputs of primary auto-coder
3. **Data structures**:
4.    $CURRENT$: map of codes to current scores
5.    $FINAL$: map of codes to final scores
6.    $QUEUE$: priority queue of scored codes
7. **For each** $D_i$ **in** $D_1 \dots D_M$:
8.    **Initialize** $CURRENT$ with $GEN(D_i)$ using primary auto-coder scores
9.    **Initialize** $QUEUE$ with $GEN(D_i)$ using primary auto-coder scores
10.    **Initialize** $FINAL$ to be empty
11.    **For** *i* **from** 1 **to** depth of exploration *d*:
12.      **Pop** $C^{top}$ **from** $QUEUE$
13.      $FINAL(C^{top}) \leftarrow CURRENT(C^{top})$
14.      **For each** $C_k$ **in** $QUEUE$:
15.        $CURRENT(C_k) \leftarrow CURRENT(C_k) \times P(C_k|C^{top})$
16.      **Update** $QUEUE$ **with** $CURRENT$
17.    **For each** $C_k$ **in** $QUEUE$:
18.      $FINAL(C_k) \leftarrow CURRENT(C_k)$
19. **Output** $FINAL$

Figure 2. Algorithm for run-time application of the model.

The only input to the algorithm is sets of top-scoring codes, $GEN(D_1) \ldots GEN(D_M)$ produced by the primary auto-coder. Its basic idea is to substitute the knowledge about what codes are appropriate with the best guess we can make based on the output of the primary auto-coder $GEN(D_i)$ for a document $D_i$. If we had to make one guess for the document, it would obviously be the code to which the primary auto-coder assigned the highest confidence score. We therefore make the approximation of assuming that this code, $C^1$, is correct and use it in place of a manually-assigned code in the model, computing the probability estimates $P(C^{pred}|C^1)$ for all other generated codes $GEN(D_1) \backslash C^1$. These probabilities reflect the propensities of the other codes to co-occur with $C^1$ We then combine these probabilities with the scores of the primary auto-coder (by simple or weighted multiplication) to produce new scores for the codes in $GEN(D_1) \backslash C^1$. To establish an iteration, we pick the code with the highest resulting score, $C^2$, and repeat the process, assuming it to be correct and using it to compute the estimates $P(C^{pred}|C^2)$ for all other codes $C^{pred}$ in $GEN(D_1) \backslash \{C^1, C^2\}$. We then multiply the previously obtained scores for the remaining codes with these probabilities, so that they now incorporate two co-occurrence estimates, which reflect co-occurrence propensities of these codes with $C^1$ and $C^2$. Repeating the iterative step up to some "depth of exploration" $d$ gives us a sequence of codes $C^1, C^2 \ldots C^d$. Whenever one of these codes has a lower propensity of co-occurring with some other code higher up in the sequence, this lowers its cumulative score and pushes it down in the ranked sequence, and vice versa.

The best guess made by this algorithm at each step may be wrong, and it could produce a chain of incorrect computations. For example, if two incompatible codes are assigned the top two scores by the primary auto-coder, with an incorrect code at the top, the algorithm would assign a low confidence to the correct code. The computational strategy of making the best guess at each step rather than looking through all possible options to find the best overall solution is known in computer science as *greedy*. It trades an increased risk of errors for higher computational efficiency. The same qualities characterize the algorithm presented here.

While the underlying code co-occurrence model could also be trained to predict correlations between procedures and diagnoses, the run-time algorithm in Figure 2 assumes that the primary auto-coder scores are directly comparable. If procedure and diagnosis codes are generated by two different systems, this may not be the case. The algorithm can be extended to handle multiple primary auto-coding systems by initializing all entries of *CURRENT* with the score of 1. That way the scores of the primary auto-coders will enter into the model only through features of the model. Their weights will be adjusted based on the training data, so that the primary auto-coder scores no longer need to be directly comparable. We show below that forgoing interpolation with the score of the primary auto-coder is a viable alternative.

**Model Features For ICD-10-PCS**

The features used by the co-occurrence model for ICD-10-PCS exploit the mapping from codes to clinical concepts, which is described in more detail in [7], as well as string patterns derived from the codes. For a pair of codes $C^{pred}$ and $C^{given}$ they include:

- A feature for the identity of the pair $C^{pred}, C^{given}$;
- A feature for the identity of $C^{pred}$ and every concept mapped to $C^{given}$;
- A feature for the identity of $C^{given}$ and every concept mapped to $C^{pred}$;
- A feature for every concept mapped to $C^{pred}$ but not to $C^{given}$;
- A feature for every concept mapped to $C^{given}$ but not to $C^{pred}$;
- A feature for every pair of concepts mapped to $C^{given}$ and $C^{pred}$, respectively;
- For codes with the same two-axis prefix, a feature encoding which axes the two codes share and on which axes they differ, using a binary representation (e.g., the pattern for codes *BP07ZZZ* and *BP08ZZZ* is *BP10111*);
- For codes with the same two-axis prefix, a feature encoding which axes the two codes share, using a binary representation, and specifying the axis values for both codes, where they differ (e.g., the pattern for codes *BP07ZZZ* and *BP08ZZZ* is *BP1{7/8}111*).

When a code-to-concept mapping is not readily available, concept-based features could be replaced by features tracking token n-grams contained in the code descriptions.

We also use features derived from the score of the primary auto-coder, including a feature for the score itself, and binary features marking whether the score exceeds specified thresholds. As will be seen below, these features play a critical role in the model.

We tested the proposed model in combination with a primary auto-coder for ICD-10-PCS described in [7], which generates codes together with probability scores based on the text of EHRs. We used a corpus of 28,044 EHRs (individual clinical records) representing a wide variety of clinical contexts. The corpus was annotated under the auspices of 3M Health Information Systems for the express purpose of developing auto-coding technology for ICD-10. Multiple coders worked on some of the documents, but they were allowed to collaborate, producing what was effectively a single set of codes for each EHR. We held out 1,001 EHRs for evaluation and used the rest for training. The same training corpus, as well as 175,798 EHRs with ICD-9 procedure codes submitted for billing by a health provider, were also used to train the models of the primary auto-coder. The top 200 predictions of the primary auto-coder were used to train the co-occurrence model. At run-time the depth of exploration was set to 3.

# RESULTS

The results of the co-occurrence models are difficult to compare against the primary auto-coder, because of the large differences in the precision/recall tradeoff between the two. Below we compare results of different experimental condition through their respective values of F-score at the decision threshold setting where precision and recall are equal (and thus equal to the F-score). The best result was obtained by interpolating the primary auto-coder score with a co-occurrence model that also used the primary auto-coder score as a feature. This condition achieved an F-score of 0.562, which represents a 12% relative improvement over the F-score of 0.501 for the primary auto-coder. To put these numbers into perspective, note that the average accuracy of trained medical coders for ICD-10 has been estimated to be 63% [12]. Using this co-occurrence model alone, without interpolating the primary auto-coder score, leads to a slight decrease in the F-score (0.547), as does omitting the features derived from the primary auto-coder score (F-score 0.542). Table 2 shows values of precision and recall for the best co-occurrence model and the primary auto-coder at roughly comparable ranges of thresholds.

| Primary auto-coder | | | Best co-occurrence model | | |
| --- | --- | --- | --- | --- | --- |
| Threshold | Precision | Recall | Threshold | Precision | Recall |
| 0.80 | 0.445 | 0.562 | 0.1 | 0.503 | 0.595 |
| 0.84 | 0.475 | 0.523 | 0.3 | 0.558 | 0.563 |
| 0.88 | 0.527 | 0.463 | 0.5 | 0.596 | 0.537 |
| 0.92 | 0.595 | 0.384 | 0.7 | 0.639 | 0.495 |
| 0.96 | 0.692 | 0.229 | 0.9 | 0.709 | 0.362 |

Table 2. Values of precision and recall at comparable ranges of thresholds for the primary auto-coder and the best co-occurrence model described in the text.

# DISCUSSION

## Related Work

The auto-coding task belongs to the class of problems called *multi-label classification* in machine learning literature, where each instance (in this case, EHR) may be assigned more than one label (in this case, medical code), in contrast to standard classification, where only one label is assigned to each instance. We assume that the primary auto-coder assigns the labels independently of each other, which has generally been the case for the machine learning methods for auto-coding proposed so far (for an overview see, e.g., Related Work in [7]), and is also common in solving multi-label classification problems in other settings [8]. Several methods for modeling inter-dependencies between labels in multi-label classification settings have also been proposed [9]. The approach we describe differs from them in several ways: it can be combined with any primary auto-coder generating confidence scores, unlike methods which model label generation from instances and interdependencies between labels at once; it is designed to handle mutually exclusive codes; it scales well to problems with a large set of labels, which is necessary for ICD-10 auto-coding; and it leverages internal label structure for better generalization from sparse training data. For illustration, consider two proposals for modeling label interdependencies that are most closely related to ours. Godbole & Sarawagi [10] propose a stacking approach where classifiers are trained to predict individual labels and their predictions are then included in the feature space of a second-level stacked classifier along with the original feature space. This resembles the stacking aspect of our model, whose features are derived both from the primary auto-coder score and from the labels it predicts. However, when the first-level classifier predicts mutually exclusive labels, their approach is not designed to select only one of them, while ours is, as shown below. Read et al [11] factor out joint probability of the labels using the chain rule into a sequence of conditional probabilities of individual labels, each conditioned on other labels, in addition to the features derived from the instance. This is somewhat similar to our conditional probability model, but their joint probability is defined over the entire inventory of labels, and they have to use an ensemble model over multiple factorizations, which would not scale for an inventory of labels as large as ICD-10. Furthermore, neither of these two papers exploits internal structure of labels, treating the labels as atomic entities instead, and they combine features derived from instances with label-based features, which would in our case duplicate the work performed by the primary auto-coder.

**Case Studies**

We illustrate the model's performance with several concrete examples. Consider the following two codes, which differ by only a single character, corresponding to procedures that are commonly performed during the same patient encounter:

- *30243**K1***: Transfusion of Nonautologous **Frozen** Plasma into Central Vein, Percutaneous Approach
- *30243**L1***: Transfusion of Nonautologous **Fresh** Plasma into Central Vein, Percutaneous Approach

Table 3 shows the probabilities assigned for combinations of these codes by the model, assuming that the primary auto-coder assigned the score of 0.9 for $C^{pred}$. These are shown in the table next to "naïve" conditional probabilities calculated from counts of the codes in the training set as

$$P(C^{pred}|C^{given}) = \frac{count(C^{pred}, C^{given})}{count(C^{given})}$$

We see that the model gives the codes similar probabilities, both of which are higher than the 0.9 assigned by the primary auto-coder. In contrast, naïve conditional probabilities show considerable variation, which appears to arise by chance.

Furthermore, we see that the model was able to make these generalizations from sparse training material, since the pair was observed in the training data only once.

Table 3 also shows the probabilities for two hip replacement codes that differ only by the device character:

- *0SR90**1**Z*: Replacement of Right Hip Joint with **Metal** Synthetic Substitute, Open Approach
- *0SR90**3**Z*: Replacement of Right Hip Joint with **Ceramic** Synthetic Substitute, Open Approach

The co-occurrence model correctly learns that the two codes are unlikely to co-occur (metal and ceramic materials cannot be implanted at once into the same joint) and generates low probabilities for both predicted codes, despite the high confidence of the primary auto-coder. Thus, whichever code has the higher primary auto-coder score, the co-occurrence model would effectively eliminate the other code from the final auto-coder output. Furthermore, unlike the naïve estimate, it achieves this goal without incorrectly eliminating valid code pairs that happen not be observed in training data by chance, which is bound to happen frequently, due to the large size of the ICD-10-PCS code set and limited training material.

| $C^{pred}$ | $C^{given}$ | Primary $C^{pred}$ score | Code pair training count | $P(C^{pred}|C^{given})$ Model | $P(C^{pred}|C^{given})$ Naïve |
|---|---|---|---|---|---|
| 30243L1 | 30243K1 | 0.9 | 1 | 0.997 | 0.167 |
| 30243K1 | 30243L1 | 0.9 | 1 | 0.998 | 1 |
| 0SR901Z | 0SR903Z | 0.9 | 0 | 0.031 | 0 |
| 0SR903Z | 0SR901Z | 0.9 | 0 | 0.013 | 0 |

Table 3. Examples of model's predictions. See text for details.

## CONCLUSION

We have presented a model of code co-occurrence propensity and a run-time algorithm for using it to rescore the output of primary auto-coders that assign clinical codes to EHRs. This approach can be used for any code set and any primary auto-coder that generates confidence scores. We have further described an application of our methods to ICD-10-PCS and showed their effectiveness.

## ACKNOWLEDGEMENTS

We would like to thank Ron Mills for providing the crosswalk from ICD-10-PCS codes to clinical concepts, as well as Roxana Safari, Jean Stoner, Lyle Schofield, Guoli

**REFERENCES**

[1] Chute GG, Yang Y, and Buntrock J. An evaluation of computer assisted clinical classification algorithms. *Proc Annu Symp Comput Appl Med Care* 1994;162-6.

[2] Heinze DT, Morsch M, Sheffer R, et al. LifeCode: A deployed application for automated medical coding. *AI Mag* 2001;2:76-88.

[3] Benson S. Computer-assisted Coding Software Improves Documentation, Coding, Compliance, and Revenue. *Perspect Health Inf Manag, Computer Assisted Coding Conference Proceedings*, Fall 2006. perspectives.ahima.org/computer-assisted-coding-software-improves-documentation-coding-compliance-and-revenue/ (accessed 9 Jul 2015)

[4] Pakhomov SV, Buntrock JD, and Chute CG. Automating the assignment of diagnosis codes to patient encounters using example-based and machine learning techniques. *J Am Med Inform Assoc* 2006;13(5):516-25.

[5] Jiang Y, Nossal M, and Resnik P. How does the system know it's right? Automated confidence assessment for compliant coding. *Perspect Health Inf Manag, Computer Assisted Coding Conference Proceedings*, Fall 2006. perspectives.ahima.org/how-does-the-system-know-its-right-automated-confidence-assessment-for-compliant-coding/ (accessed 9 Jul 2015)

[6] Gao G, Andrew G, Johnson M, et al. A comparative study of parameter estimation methods for statistical natural language processing. *Proceedings of the 45th Annual Meeting of the Association for Computational Linguistics* 2007; 824-831.

[7] Subotin M and Davis AR. A system for predicting ICD-10-PCS codes from electronic health records. *Proceedings of BioNLP* 2014;59-67.

[8] Tsoumakas G and Katakis I. Multi-label classification: an overview. *International Journal of Data Warehousing and Mining* 2007;3:1-13.

[9] Dembczyński K, Waegeman W, Cheng W, et al. On label dependence and loss minimization in multi-label classification. *Machine Learning* 2012:88: 5-45.

[10] Godbole S and Sarawagi S. Discriminative methods for multi-labeled classification. *Advances in Knowledge Discovery and Data Mining, Lecture Notes in Computer Science* 2004;3056:22-30.

[11] Read J, Pfahringer B, Holmes G, et al. Classifier chains for multi-label classification. *Machine Learning* 2011;85:333-359.

[12] HIMSS/WEDI. ICD-10 National Pilot Program Outcomes Report. 2013. files.himss.org/FileDownloads/ICD-10_NPP_Outcomes_Report.pdf (accessed 9 Jul 2015)